\documentclass{article}

\usepackage{arxiv}

\usepackage[utf8]{inputenc} 
\usepackage[T1]{fontenc}    
\usepackage{url}            
\usepackage{booktabs}       
\usepackage{nicefrac}       
\usepackage{microtype}      
\usepackage{lipsum}
\usepackage{graphicx}

\graphicspath{{./figures/}}
\usepackage[export]{adjustbox}
\usepackage{amsmath,amssymb,amsthm,amsfonts}
\usepackage{mathrsfs}
\usepackage{scalerel}
\usepackage{float}
\usepackage{caption}
\usepackage{multirow}
\usepackage{centernot}
\usepackage{mathtools}
\usepackage{algcompatible}
\usepackage[algoruled,linesnumbered,vlined,inoutnumbered]{algorithm2e}
\usepackage{subcaption}
\usepackage[hidelinks]{hyperref}       
\usepackage{url}            
\usepackage[dvipsnames]{xcolor}
\usepackage{cleveref}

\newcommand{\beq}{\begin{equation}}
\newcommand{\eeq}{\end{equation}}

\let\oldnl\nl
\newcommand{\nonl}{\renewcommand{\nl}{\let\nl\oldnl}}

\newcommand{\appropto}{\mathrel{\vcenter{
  \offinterlineskip\halign{\hfil$##$\cr
    \propto\cr\noalign{\kern2pt}\sim\cr\noalign{\kern-2pt}}}}}

\title{Filtered Manifold Alignment}

\author{
Stefan Dernbach \\
  College of Information and Computer Sciences\\
  University of Massachusetts\\
  Amherst, MA 01003 \\
  \texttt{dernbach@cs.umass.edu} \\
   \And
Don Towsley \\
  College of Information and Computer Sciences\\
  University of Massachusetts\\
  Amherst, MA 01003 \\
  \texttt{towsley@cs.umass.edu} \\
}

\begin{document}
\maketitle
\begin{abstract}
Domain adaptation is an essential task in transfer learning to leverage data in one domain to bolster learning in another domain. In this paper, we present a new semi-supervised manifold alignment technique based on a two-step approach of projecting and filtering the source and target domains to low dimensional spaces followed by joining the two spaces. Our proposed approach, filtered manifold alignment (FMA), reduces the computational complexity of previous manifold alignment techniques, is flexible enough to align domains with completely disparate sets of feature and demonstrates state-of-the-art classification accuracy on multiple benchmark domain adaptation tasks composed of classifying real world image datasets.
\end{abstract}


\section{Introduction}
In many domains, raw data has become abundant and inexpensive, while labeled data often remains sparse and costly to create. Leveraging available labeled data to bolster learning on related unlabeled or sparsely labeled datasets is essential to cost-effective machine learning. Transfer learning is the process of adapting knowledge from one problem to another. Domain adaptation is a subset of transfer learning focused on applying knowledge of a specific problem from one domain, the source, to a second domain, the target. The data distribution of the source domain often varies from that of the target domain. For example, the pose, lighting, and backgrounds of a set of images from an amateur photographer may differ significantly from precompiled, labeled photos taken by a professional. These differences can cause a significant degradation of the accuracy of a machine learning model trained on one dataset and applied to another when they are not adjusted for. 

Manifold alignment (MA) is a domain adaptation approach based on the theory that data in both the source and target domains are drawn from the same underlying, low-dimensional manifold. MA projects and aligns two or more datasets onto this manifold so that learning can be done in the joint space. This work introduces a new semi-supervised filtered manifold alignment (FMA) technique in which we align two datasets by first learning an individual embedding for each domain based on a discrete graph approximation of the underlying manifold and then subsequently aligning these two embeddings by connecting the two graphs via cross-domain correspondences. The initial embeddings filter noise in the original domains prior to aligning the two domains. This ensures that the filter of one domain does not affect the second domain or the cross-domain correspondences. This approach offers low computational complexity, can be applied to new samples even after the alignment is complete, and is applicable to heterogeneous domains (i.e. domains with different feature sets). Additionally, a feature-based version of our method further simplifies the task of embedding new data points that weren't part of the original alignment. The complexity of the feature-based method scales with the number of features rather than the number of samples in the dataset allowing one to choose the method most suited for their task.

\section{Manifold Alignment}
\label{sec:MA}
Manifold alignment~\cite{wang2011manifold} facilitates knowledge transfer between two domains by utilizing correspondences across the domains to align their underlying manifolds. Semi-supervised manifold alignment (SMA)~\cite{ham2005semisupervised} combines the task of projecting data down to its underlying low-dimensional manifold and aligning instances across datasets. Given datasets $\mathbf{X}^{(1)}\in\mathbb{R}^{m_1\times n_1}$ and $\mathbf{X}^{(2)}\in\mathbb{R}^{m_2\times n_2}$ with possibly different numbers of samples, $m_1\neq m_2$, and features, $n_1\neq n_2$, SMA seeks low $n$-dimensional embeddings $\mathbf{Z}^{(1)}\in\mathbb{R}^{m_1\times n}$ and $\mathbf{Z}^{(2)}\in\mathbb{R}^{m_1\times n}$ that preserve intra-dataset relationships and inter-dataset correspondences. The intra-dataset relationships can be calculated using a similarity measure, $sim(a,b)$, between two samples such as the cosine similarity of their feature vectors. The inter-dataset correspondences, $cor(a,b)$, are determined by matching a small set of labeled instances from the target domain with labeled instances in the source domain. Weight matrices $\{\mathbf{W}^{(1)},\mathbf{W}^{(2)},\mathbf{W}^*\}$, where $\mathbf{W}^{(a)}_{ij}=sim\left(\mathbf{X}^{(a)}_i,\mathbf{X}^{(a)}_j\right)$ encodes the similarity (e.g. cos similarity) between samples $i$ and $j$ in dataset $a\in\{1,2\}$ and $\mathbf{W}^*_{ij}=cor\left(\mathbf{X}^{(1)}_i,\mathbf{X}^{(2)}_j\right)$ encodes the correspondence between sample $i$ in the source domain and sample $j$ in the target domain (e.g. $1$ when they share a label and $0$ if the labels differ or are unknown). In an optimization setting, the two goals of preserving intra- and inter-dataset relationships are expressed as a pair of loss functions:
\begin{align}
L_1\left(Z\right) &= \sum_{a=1}^2\sum_{i,j} \mathbf{W}^{(a)}_{ij}||Z^{(a)}_i-Z^{(a)}_j||^2_{l^2}
\label{eq:l1}\\
    L_2\left(Z\right) &= \sum_{i,j} \mathbf{W}^*_{ij}||Z^{(1)}_i-Z^{(2)}_j||^2_{l^2}.
    \label{eq:l2}
\end{align}
The loss in \eqref{eq:l1} penalizes the embedding distance between related examples within a dataset and \eqref{eq:l2} penalizes the embedding distance between corresponding examples across the two datasets. Combining the matrices such that $\mathbf{Z}=\begin{bmatrix} \mathbf{Z}^{(1)} & \mathbf{0} \\ \mathbf{0} & \mathbf{Z}^{(2)} \end{bmatrix}$ and $\mathbf{W} = \begin{bmatrix} \mathbf{W}^{(1)} & \mathbf{W}^* \\ \mathbf{W}^* & \mathbf{W}^2 \end{bmatrix}$ and summing the two loss functions produces the combined loss:
\begin{align}
L(\mathbf{Z}) &=\sum_{i,j}\mathbf{W}(i,j)||\mathbf{Z}[i]-\mathbf{Z}[j]||^2_{l^2} \\
\label{eq:trMA}
&= tr(\mathbf{Z}^T\mathbf{L}\mathbf{Z}),    
\end{align} 
where  $\mathbf{L}=\begin{bmatrix}\mathbf{D}^{(1)}-\mathbf{W}^{(1)} & -\mathbf{W}^* \\ -\mathbf{W}^* & \mathbf{D}^{(2)}-\mathbf{W}^{(2)}\end{bmatrix}$ is referred to as the joint Laplacian. Given the joint degree matrix $\mathbf{D}=\begin{bmatrix}\mathbf{D}^{(1)} & \mathbf{0} \\ \mathbf{0} & \mathbf{D}^{(2)}\end{bmatrix}$, the constraint $\mathbf{Z}^T\mathbf{D}\mathbf{Z}=\mathbf{I}$ removes trivial solutions to the optimization problem. Minimizing~\eqref{eq:trMA}, subject to this constraint, is equivalent to solving the generalized eigenvalue problem for the first $n$ nontrivial eigenvectors $\pmb\Phi$ corresponding to the smallest eigenvalues $\pmb\Lambda$: 
\begin{equation}
\mathbf{L}\pmb\Phi= \mathbf{D}\pmb\Phi\pmb\Lambda. \label{eq:MA}
\end{equation}
The rows of the eigenvector matrix provide low dimensional embeddings of each sample, $\mathbf{X}_i\rightarrow\pmb\Phi_i$, that allows samples from different domains to be compared.

\section{Filtered Manifold Alignment}
\label{sec:FMA}
The approach in filtered manifold alignment (FMA) is to separate the process of embedding each dataset into a low dimensional space from the process of aligning the embeddings. FMA calculates the spectra of the graph Laplacians associated with the source and target domains individually and combines them into an approximation of the spectra of the joint graph Laplacian.

In order to separate the embedding process from the alignment process, FMA first converts the generalized eigenvector equation in ~\eqref{eq:MA} into a standard eigenvector problem. Because $\mathbf{D}$ is a diagonal matrix, this conversion is straightforward and produces the standard eigenproblem:
\begin{equation}
\label{eq:fma}
   \mathbf{D}^{-1/2} \mathbf{L} \mathbf{D}^{-1/2}\pmb\Phi = \pmb\Phi \pmb \Lambda.
\end{equation} 
The $n$-dimensional embeddings of each sample for semisupervised manifold alignment are now given by the rows of the matrix $\mathbf{Z}=\mathbf{D}^{-1/2}\pmb\Phi_{*,1:n}\pmb\Lambda^{-1/2}_{1:n,1:n}$. 

Once converted into a standard eigenproblem, the next step is to divide the embedding and aligning tasks. This is done by separating the joint Laplacian into two matrices. The joint graph Laplacian is the sum of the disconnected Laplacian $\mathbf{L}^*=\begin{bmatrix}\mathbf{L}^{(1)} & \mathbf{0} \\ \mathbf{0} & \mathbf{L}^{(2)} \end{bmatrix}$ and the product of the cross-domain incidence matrix with its transpose: $\mathbf{L}=\mathbf{L}^*+\mathbf{A}\mathbf{A}^T$. The cross-domain incidence matrix $\mathbf{A}$ maps each cross-domain correspondence pair $(i,j)$ to a unique column $k$ in $\mathbf{A}$, such that $\mathbf{A}_{ik}=1$ and $\mathbf{A}_{jk}=-1$. All other elements of $\mathbf{A}$ are $0$. The L.H.S. of~\eqref{eq:fma} becomes: 
\begin{equation}
    \mathbf{D}^{-1/2}\mathbf{L}\mathbf{D}^{-1/2}\pmb\Phi=\left(\mathbf{D}^{-1/2}\mathbf{L}^*\mathbf{D}^{-1/2}+\mathbf{D}^{-1/2}\mathbf{A}\mathbf{A}^T\mathbf{D}^{-1/2}\right)\pmb\Phi.
\end{equation} 
This eigenproblem is solved by determining the eigenvectors and eigenvalues of $\mathbf{D}^{-1/2}\mathbf{L}^*\mathbf{D}^{-1/2}$ and treating $\mathbf{D}^{-1/2}\mathbf{A}\mathbf{A}^T\mathbf{D}^{-1/2}$ as an update matrix to learn the eigenvectors and eigenvalues of $\mathbf{D}^{-1/2}\mathbf{L}\mathbf{D}^{-1/2}$.

\begin{figure}
  \centering
  \begin{subfigure}[t]{0.32\linewidth}
  \centering
  \includegraphics[width=\linewidth,frame]{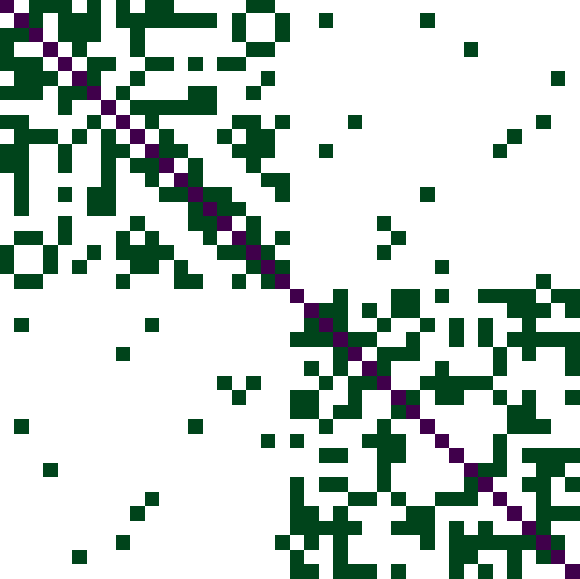}
  \caption{Original Laplacian}
  \label{L}
  \end{subfigure}
  \begin{subfigure}[t]{0.32\linewidth}
  \centering
  \includegraphics[width=\linewidth,frame]{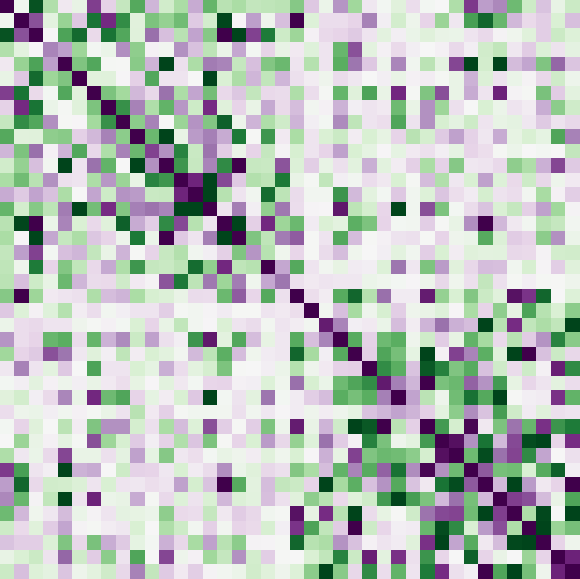}
  \caption{Standard Filter}
  \label{MA}
  \end{subfigure}  
  \begin{subfigure}[t]{0.32\linewidth}
  \centering
  \includegraphics[width=\linewidth,frame]{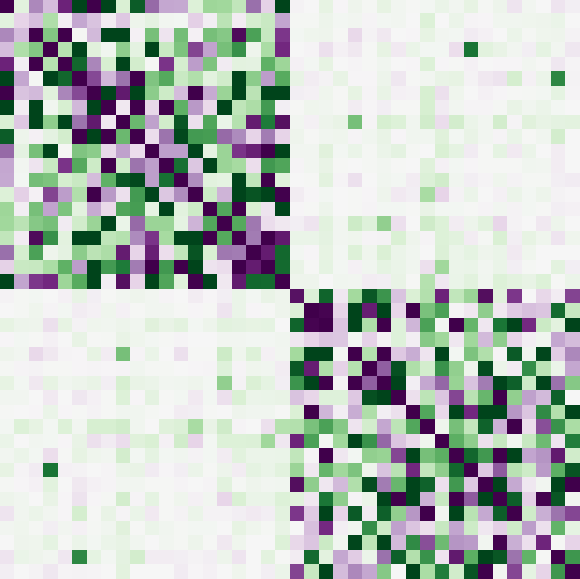}
  \caption{Subgraph Filter}
    \label{fma}
  \end{subfigure}
  
  \caption{\textbf{Comparison of graph filtering methods}. The original Laplacian (\subref{L}) is composed of two 20 node Erdős–Rényi graphs with several random connecting edges. The standard filter (\subref{MA}) shows the matrix recomposed from half of its eigenvectors. The subgraph filter (\subref{fma}) performs the same eigen-filter on each subgraph before combining them via the linking edges resulting in smoother off-diagonal blocks.}
      \label{fig:filter}
\end{figure}

Because the disconnected Laplacian is block diagonal and symmetric (each block being the Laplacian of one domain), the eigendecomposition of $\mathbf{D}^{-1/2}\mathbf{L}^*\mathbf{D}^{-1/2}$ can be efficiently computed by computing the eigendecomposition of each block and then combining them:
\begin{equation}\label{eq:djl}
    \mathbf{D}^{-1/2}\mathbf{L}^*\mathbf{D}^{-1/2}=\begin{bmatrix}\pmb\Phi^{(1)} & \mathbf{0} \\ \mathbf{0} & \pmb\Phi^{(2)} \end{bmatrix} \begin{bmatrix}\pmb\Lambda^{(1)} & \mathbf{0} \\ \mathbf{0} & \pmb\Lambda^{(2)} \end{bmatrix} \begin{bmatrix}\pmb\Phi^{(1)} & \mathbf{0} \\ \mathbf{0} & \pmb\Phi^{(2)}\end{bmatrix}^T.
\end{equation}
FMA retains the smallest $n/2$ eigenvectors of each block in~\eqref{eq:djl}. The eigendecomposition process of each block is equivalent to using Laplacian eigenmaps~\cite{belkin2003laplacian} to embed each of the original datasets. The rows of the matrix $\pmb\Phi^{(i)}$ produces an $n$-dimensional embedding of the original samples in $\mathbf{X}^{(i)}$ that best preserves the local geometry, according to the loss in~\eqref{eq:l1}. This subspace embedding has the effect of denoising the manifold represented by the graphs of each domain by filtering out eigenvector-eigenvalue pairs associated with the highest frequencies on the graph~\cite{deutsch2016manifold}. By minimizing~\eqref{eq:l1} separately from~\eqref{eq:l2}, this filtering focusses on noise in the intra-dataset relationships and not the cross domain correspondences. This filtering effect can be seen in Figure~\ref{fig:filter} where the Joint Laplacian of two graphs is decomposed, filtered, and reassembled.

\begin{algorithm}
    \caption{SVD Update}
    \label{alg:iSVD}
    \begin{algorithmic}[1]
        \STATEx Solves $\mathbf{U}'\mathbf{S}'\mathbf{V'}^T=\mathbf{X}+\mathbf{AB}^T$ given $\mathbf{U}\mathbf{S}\mathbf{V}^T=\mathbf{X}$ 
        \STATE \textbf{Input:} $\mathbf{U}$, $\mathbf{S}$, $\mathbf{V}$, $\mathbf{A}$, $\mathbf{B}$
        \STATE $\begin{bmatrix}\mathbf{U} & \mathbf{P}\end{bmatrix}
                \begin{bmatrix}\mathbf{I} & \mathbf{U}^T\mathbf{A} \\ \mathbf{0} & \mathbf{R}_A\end{bmatrix} \leftarrow  QR\left(\begin{bmatrix}\mathbf{U} & \mathbf{A}\end{bmatrix}\right)$
        \STATE $\begin{bmatrix}\mathbf{V} & \mathbf{Q}\end{bmatrix}
                \begin{bmatrix}\mathbf{I} & \mathbf{V}^T\mathbf{B} \\ \mathbf{0} & \mathbf{R}_B\end{bmatrix} \leftarrow QR \left( \begin{bmatrix}\mathbf{V} & \mathbf{B}\end{bmatrix}\right)$
        
        \STATE $\mathbf{Z}=\begin{bmatrix}\mathbf{S} & \mathbf{0} \\ \mathbf{0} & \mathbf{0} \end{bmatrix}-
                \begin{bmatrix}\mathbf{U}^T\mathbf{A} \\\mathbf{R}_A\end{bmatrix}
                \begin{bmatrix}\mathbf{V}^T\mathbf{B} \\ \mathbf{R}_B\end{bmatrix}^T$
        \STATE $\mathbf{U}^*,\mathbf{S}^*,\mathbf{V}^* \leftarrow svd(\mathbf{Z})$
        \STATE $\hat{\mathbf{U}}\leftarrow \begin{bmatrix}\mathbf{U} & \mathbf{P}\end{bmatrix}\mathbf{U}^*$
        \STATE $\hat{\mathbf{S}}=\mathbf{S}^*$
        \STATE $\hat{\mathbf{V}}\leftarrow \begin{bmatrix}\mathbf{V} & \mathbf{Q}\end{bmatrix}\mathbf{V}^*$
        \STATE \textbf{Return:} $\hat{\mathbf{U}},\hat{\mathbf{S}},\hat{\mathbf{V}}$
    \end{algorithmic}
\end{algorithm}

\begin{algorithm}
    \caption{Block SVD Update}
    \label{alg:bSVDu}
    \begin{algorithmic}[1]
    \STATEx Updates the eigenvectors and eigenvalues for $\mathbf{X} + \mathbf{A}\mathbf{A}^T$ given $\mathbf{X} = \pmb\Phi\pmb\Lambda\pmb\Phi^T$
        \STATE \textbf{Input:} $\pmb\Phi$, $\pmb\Lambda$, $\mathbf{A}$
        \STATE $\pmb\Phi',\pmb\Lambda',\pmb\Phi'^{T}\leftarrow svd\left(\pmb\Lambda+\pmb\Phi^T\mathbf{A}\mathbf{A}^T\pmb\Phi\right)$
        \STATE $\pmb\Phi''\leftarrow \pmb\Phi\pmb\Phi'$
        \STATE \textbf{Return:} $\pmb\Phi'',\pmb\Lambda'$
    \end{algorithmic}
\end{algorithm}

The second step in FMA is to align the projections for each dataset by updating the joint eigenvectors and eigenvalues using the cross-domain correspondences. We adapt the low-rank spectral update in 
\cite{brand2003fast}. Given $\mathbf{U}\mathbf{S}\mathbf{V}^T=\mathbf{X}$, the spectral update (Algorithm~\ref{alg:iSVD}) computes new singular values and vectors $\hat{\mathbf{U}}\hat{\mathbf{S}}\hat{\mathbf{V}}^T = \mathbf{X}+\mathbf{A}\mathbf{B}^T$ without ever directly referencing $\mathbf{X}$. We apply the update algorithm to the eigenvectors and eigenvalues computed above. 

Alg.~\ref{alg:bSVDu} reduces the complexity of the spectral update under the following assumptions: $\mathbf{X}$ is symmetric, $\mathbf{A}=\mathbf{B}$, and only the projection of $\mathbf{A}$ onto the span of $\mathbf{X}$ is relevant. These assumptions hold when updating the eigenvectors of $\mathbf{L}^*+\mathbf{A}\mathbf{A}^T$. The last condition, specifically, is met by FMA because only the smallest eigenvalues and eigenvectors are used in the embedding. Any information in $\mathbf{A}$ that lies outside of the smallest eigenvectors of $\mathbf{L}^*$ will only increase the eigenvalues of the corresponding vectors and thus is likely to continue to be filtered out. Alg.~\ref{alg:bSVDu} is efficient when $rank(\mathbf{X})$ is small because it relies on a spectral decomposition of a square matrix whose size is equal to its rank. By performing the filtering in the first step of FMA, the rank of the disconnected Laplacian has been reduced, making the update step efficient. The complete FMA algorithm to embed two datasets is given in Alg. \ref{alg:FMA}.

\begin{figure}
  \centering
  \begin{subfigure}[t]{0.49\linewidth}
  \centering
  \includegraphics[width=\linewidth]{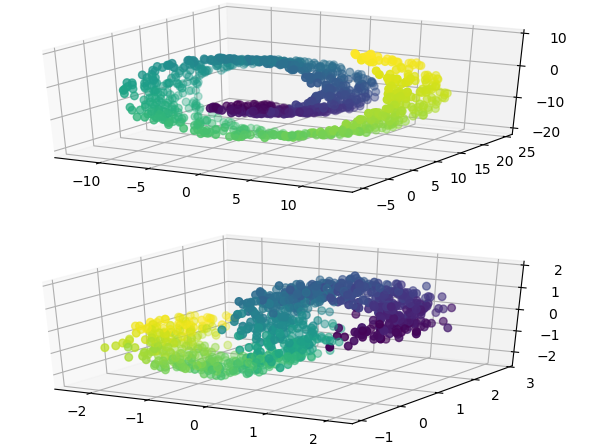}
  \caption{Original Data Points}
  \label{swissS}
  \end{subfigure}
  \begin{subfigure}[t]{0.49\linewidth}
  \centering
  \includegraphics[width=\linewidth]{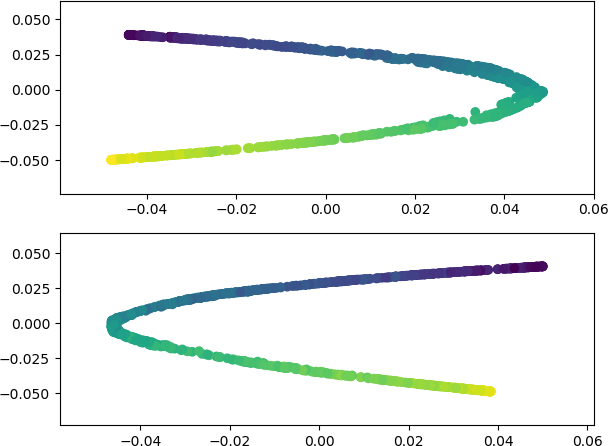}
  \caption{Independent Spectral Embeddings}
  \label{indemb}
  \end{subfigure}  \\
  \begin{subfigure}[t]{0.48\linewidth}
  \centering
  \includegraphics[width=\linewidth]{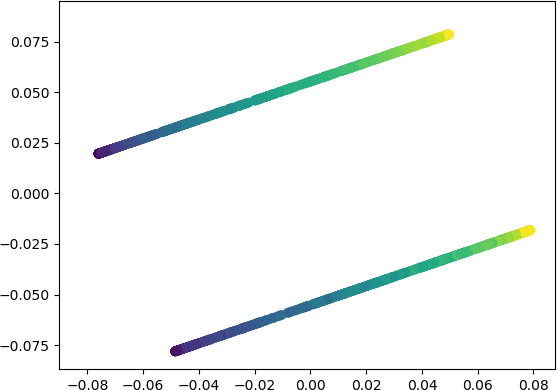}
  \caption{Joint Embedding}
  \label{jointemb}
  \end{subfigure}
\caption{\textbf{Example FMA process.} (\subref{swissS}) A random sample of 400 points are collected from a noisy 3D swiss roll manifold and a noisy 3D S-curve manifold. (\subref{indemb}) The datapoints from the two manifold are embedded independently onto a two dimensional manifold via spectral embedding. (\subref{jointemb}) The two embeddings are aligned onto the same 1D manifold.}
\label{fmademo}
\end{figure}

An illustrative example of the process FMA performs is given in Figure~\ref{fmademo}. The two datasets being aligned are composed of 400 points sampled from two different three-dimensional manifolds: the swiss roll manifold and the S-curve manifold (Figure \ref{swissS}). Noise is added to the points and they are colored according to their location along the true intrinsic 1 dimensional manifold. A nearest neighbor graph, using the 5 nearest neighbors of each point is formed for each of the two datasets. Each point is embedded onto the first two nontrivial eigenvectors of the Laplacian matrix of the nearest neighbor graphs (Figure~\ref{indemb}). The joint graph is formed using $40$ corresponding points across the two manifolds and the block SVD update is used to combine and align the two embeddings and reduce the embeddings to 1D (Figure~\ref{jointemb}). 

\subsection{Feature-Level Alignment}
Semi-supervised manifold alignment can be modified to find a linear mapping from the original feature spaces to the embedding space. This is referred to as linear or feature-level alignment in contrast to nonlinear instance-level alignment because it maps features of the original domains into the joint embedding space rather than mapping instances directly. Replacing~\eqref{eq:MA} with the new generalized eigenproblem~\cite{wang2009general}:
\begin{equation}
    \mathbf{X}^T\mathbf{L}\mathbf{X}\pmb\Phi=\mathbf{X}^T\mathbf{D}\mathbf{X}\pmb\Phi\pmb\Lambda, \label{eq:lMA}
\end{equation}
allows us to solve for the eigenvectors $\pmb\Phi$, which can be used to map features from the original domains into the joint embedding space $\mathbf{X}\rightarrow \mathbf{X}\pmb\Phi$. The R.H.S of~\eqref{eq:lMA} now contains the dense matrix $\mathbf{X}$ in addition to the diagonal matrix $\mathbf{D}$ in~\eqref{eq:MA}. The conversion to a regular eigenvector problem requires an extra step to calculate $(\mathbf{X}^T\mathbf{D}\mathbf{X})^{-1/2}$. The adjusted algorithm is given in Alg.~\ref{alg:FFMA}.

\begin{algorithm}
    \caption{Filtered Manifold Alignment}
    \label{alg:FMA}
    \begin{algorithmic}[1]
        \STATE \textbf{Input:} Data Matrices: $\mathbf{X}^{(1)}\in \mathbb{R}^{m_1 \times n_1}$, $\mathbf{X}^{(2)}\in \mathbb{R}^{m_2 \times n_2}$
        \STATEx \hspace{29pt} Cross Correspondence Incidence Matrix: $\mathbf{A}$
        \STATEx \hspace{29pt} Neighbors: $k$
        \STATEx \hspace{29pt} Embedding Dimension: $n$
        \FORALL{$\mathbf{X}^{(i)}$}
            \STATE $\mathbf{L}^{(i)} \leftarrow knn(\mathbf{X}^{(i)},k)$
            \STATE $\pmb\Phi^{(i)},\pmb\Lambda^{(i)} \leftarrow eig_{n/2}\left((\mathbf{D}^{(i)})^{-1/2}\mathbf{L}^{(i)}(\mathbf{D}^{(i)})^{-1/2}\right)$
        \ENDFOR
        \STATE $\pmb\Phi,\pmb\Lambda\leftarrow\begin{bmatrix}\pmb\Phi^{(1)} & \mathbf{0} \\ \mathbf{0} & \pmb\Phi^{(2)} \end{bmatrix}, \begin{bmatrix}\pmb\Lambda^{(1)} & \mathbf{0} \\ \mathbf{0} & \pmb\Lambda^{(2)} \end{bmatrix}$
        \STATE $\mathbf{A}'\leftarrow\mathbf{D}^{-1/2}\mathbf{A}$
        \STATE $\pmb\Phi',\pmb\Lambda' \leftarrow svdu(\pmb\Phi,\pmb\Lambda,\mathbf{A}')$ \tcp*{Alg. \ref{alg:bSVDu}}
        \STATE $\mathbf{Z} \leftarrow \mathbf{D}^{-1/2} \pmb\Phi' \pmb\Lambda'^{-1/2}$
        \STATE \textbf{Return:} $\mathbf{Z}_{1:m_1,1:n}$, $\mathbf{Z}_{m_1+1:m_1+m_2,1:n}$
    \end{algorithmic}
\end{algorithm}

\begin{algorithm}
    \caption{Linear Filtered Manifold Alignment}
    \label{alg:FFMA}
    \begin{algorithmic}[1]
        \STATE \textbf{Input:} Data Matrices: $\mathbf{X}^{(1)}\in \mathbb{R}^{m_1 \times n_1}$, $\mathbf{X}^{(2)}\in \mathbb{R}^{m_2 \times n_2}$
        \STATEx \hspace{32pt} Cross Correspondence Incidence Matrix: $\mathbf{A}$
        \STATEx \hspace{32pt} Neighbors: $k$
        \STATEx \hspace{32pt} Embedding Dimension: $n$
        \FORALL{$\mathbf{X}^{(i)}$}
            \STATE $\mathbf{L}^{(i)},\mathbf{D}^{(i)} \leftarrow knn(\mathbf{X}^{(i)},k)$
            \STATE $\mathbf{T}=\left((\mathbf{X}^{(i)})^T\mathbf{D}^{(i)}\mathbf{X}^{(i)}\right)^{-1/2}$ 
            \STATE $\pmb\Phi^{(i)},\pmb\Lambda^{(i)} \leftarrow eig_{n/2}\left((\mathbf{X}^{(i)}\mathbf{T}^{(i)})^T\mathbf{L}^{(i)}\mathbf{X}^{(i)}\mathbf{T}^{(i)}\right)$
        \ENDFOR
        \STATE $\pmb\Phi,\pmb\Lambda\leftarrow \begin{bmatrix}\pmb\Phi^{(1)} & \mathbf{0} \\ \mathbf{0} & \pmb\Phi^{(2)} \end{bmatrix}, \begin{bmatrix}\pmb\Lambda^{(1)} & \mathbf{0} \\ \mathbf{0} & \pmb\Lambda^{(2)} \end{bmatrix}$
        \STATE $\mathbf{A}'\leftarrow \mathbf{A}\mathbf{T}$
        \STATE $\pmb\Phi',\pmb\Lambda'\leftarrow bsvdu(\pmb\Phi,\pmb\Lambda,\mathbf{A}')$ \tcp*{Alg. \ref{alg:bSVDu}}
        \STATE $\mathbf{Z} \leftarrow \mathbf{X} \mathbf{T} \pmb\Phi'(\pmb\Lambda')^{-1/2}$
        \STATE \textbf{Return:} $\mathbf{Z}_{1:n_1,1:n}$, $\mathbf{Z}_{n_1+1:n_1+n_2,1:n}$
    \end{algorithmic}
\end{algorithm}

\subsection{Complexity Analysis and Extensions}
Filtered manifold alignment computation is dominated by spectral decompositions of several matrices. These include decompositions of each of the sparse graph Laplacian matrices, each of which has a naïve complexity of $O(N^3)$, and a decomposition of an $n\times n$ matrix where $n\ll N$ is the dimension of the final embedding space. The summed complexity of three smaller singular value decompositions is a large improvement over single step semi-supervised manifold alignment which diagonalizes the joint Laplacian matrix $O((2N)^3)$. Taking advantage of sparse solvers can further speed up both methods. Feature-level filtered manifold alignment scales with the size of the feature space rather than the number of samples. The spectral decompositions of the graph Laplacian matrices are replaced with $M\times M$ matrices where $M$ is the number of features leading to a complexity of $O(M^3)$. As the number of samples in a dataset grows in comparison to the number of features per sample, linear FMA becomes more efficient compared to nonlinear FMA.

Feature-level filtered manifold alignment can be trivially applied to samples not in the initial alignment because it learns a linear mapping from the feature space of a sample to the joint embedding space. Nonlinear alignment, which learns a direct map for each sample, can not be directly applied to new samples. However, the spectral update (Alg.~\ref{alg:bSVDu}) can be applied to embed a new sample using the incidence matrix of the nearest neighbors as the update matrix. Additionally, both instance-level and feature-level schemes can align more than 2 datasets at a time by extending the block diagonal formatting of the joint Laplacian and related matrices.

\section{Related Work}
Domain Adaptation is a large area of study that stretches across disciplines such as computer vision, natural language processing, and machine learning. It is studied as an unsupervised, semi-supervised, or supervised problem. We focus on reviewing unsupervised and semi-supervised methods below.

Geodesic Flow Kernel (GFK)~\cite{gong2012geodesic} is an unsupervised domain adaptation method. GFK treats the source domain and the target domain as points on a Grassmanian manifold defined by their principal components. An infinite-dimensional feature space, $\mathcal{H}^{\infty}$, can be constructed by interpolating between the source and target points. An inner product can be computed in $\mathcal{H}^{\infty}$ to construct a kernelized classifier: $\left <z_i,z_j \right>=x_i^T \mathbf{G} x_j$, where $z_i$ are the transformed features of sample point $x_i$ and $\mathbf{G}$ is the kernel matrix.

Manifold Embedded Distribution Alignment (MEDA)~\cite{wang2018visual}, like FMA, takes a two step approach to dataset alignment. MEDA first performs manifold feature learning on each domain followed by dynamic distribution alignment. The manifold feature learning is done using GFK where learned features $z_i=\sqrt{\mathbf{G}x_i}$. The second step, dynamic distribution alignment, balances the importance of the marginal distribution and conditional distributions of samples between the domains. Soft labels, using a classifier on $z_i$ are used in place of actual labels in the target domain when calculating the distributions.

Correlation Alignment (CORAL)~\cite{sun2016return} transforms the source domain to match the target domain by first whitening the domain using the inverse square root of the covariance matrix of the source domain and then recoloring using the target covariance matrix: $Z_S = X_S\mathbf{C}_S^{-1/2}\mathbf{C}_T^{1/2}$, where $\mathbf{C}_i=cov(X_i)+\mathbf{I}$.

Semi-Supervised Subspace Alignment (SSA)~\cite{yao2015semi} learns orthogonal linear maps for source and target domains by simultaneously preserving intra-domain and inter-domain relationships and reducing classification error. CORAL does so by minimizing a loss function that is a weighted sum of the empirical risk, the distance between corresponding labeled points across the two domains, and a manifold regularizer on the pairwise similarities of the unlabeled samples in the target domain.

Manifold alignment methods~\cite{ham2005semisupervised,wang2008manifold,wang2011heterogeneous} are nonlinear embedding methods that work by forming a joining graph between samples within and across domains. The eigenvectors of the Laplacian of this graph corresponding to the smallest eigenvalues embed the original domains in a way to preserve the embedded distance between points.

\section{Experiments}
\label{sec:Exp}
In this section, we perform multiple experiments to illustrate the effectiveness of nonlinear instance-level filtered manifold alignment (FMA-I) and linear, feature-level filtered manifold alignment (FMA-F). We compare the performance of both filtered manifold alignment approaches to that of several state-of-the-art domain adaptation methods: Geodesic Flow Kernel (GFK)~\cite{gong2012geodesic}, Manifold Embedded Distribution Alignment (MEDA)~\cite{wang2018visual}, Correlation Alignment (CORAL)~\cite{sun2016return}, Semi-Supervised Subspace Alignment (SSA)~\cite{yao2015semi}, and Semi-Supervised Manifold Alignment (SMA)~\cite{ham2005semisupervised}. GFK, MEDA, and CORAL are unsupervised methods. SSA and SMA  are semi-supervised.

\subsection{Datasets}
We use 9 different datasets organized into 3 different experimental groups based upon shared classes between the datasets.\footnote{Datasets available at \url{https://github.com/jindongwang/transferlearning}} Each dataset is composed of a collection of images for which SURF (speeded up robust features)~\cite{saenko2010adapting} and/or DeCaf6~\cite{donahue2014decaf} features are provided. The DeCaf6 features are the hidden representations of the images extracted from the second to last layer of the deep DeCaf convolutional neural network. In general, Decaf6 features are more descriptive than SURF features and provide better classification accuracy. The datasets were chosen because they are popular benchmarks composed of real world images.

\begin{figure}
  \centering
  \begin{subfigure}[t]{0.49\linewidth}
  \centering
  \includegraphics[width=0.23\linewidth]{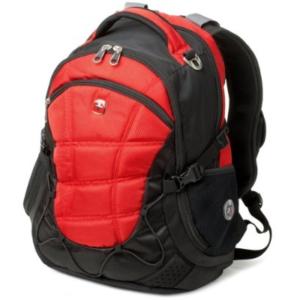}
  \includegraphics[width=0.23\linewidth]{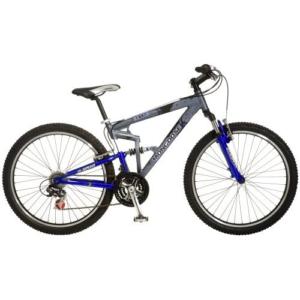}
  \includegraphics[width=0.23\linewidth]{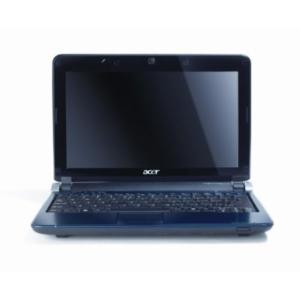}
  \includegraphics[width=0.23\linewidth]{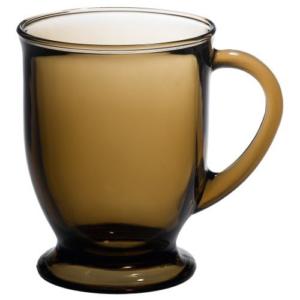}
  \caption{Amazon}
  \label{amzn}
  \end{subfigure}
  \begin{subfigure}[t]{0.49\linewidth}
  \centering
  \includegraphics[width=0.23\linewidth]{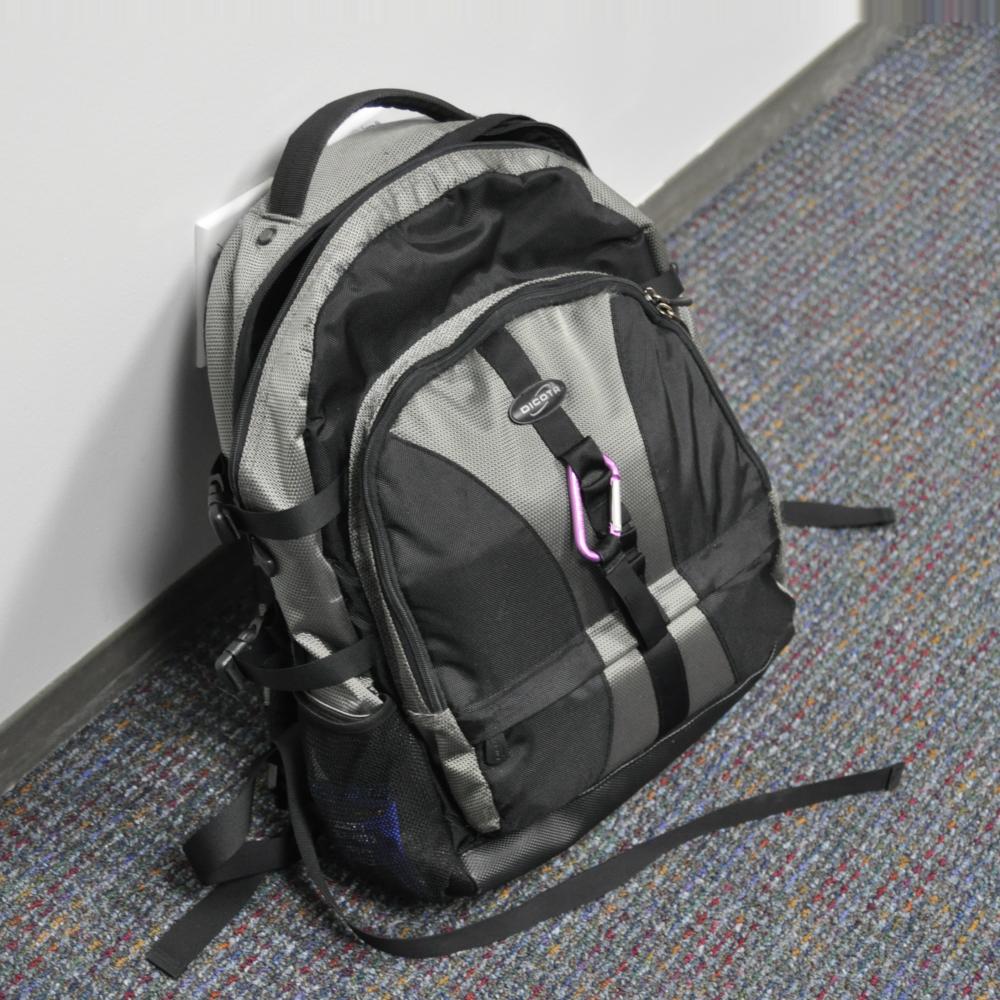}
  \includegraphics[width=0.23\linewidth]{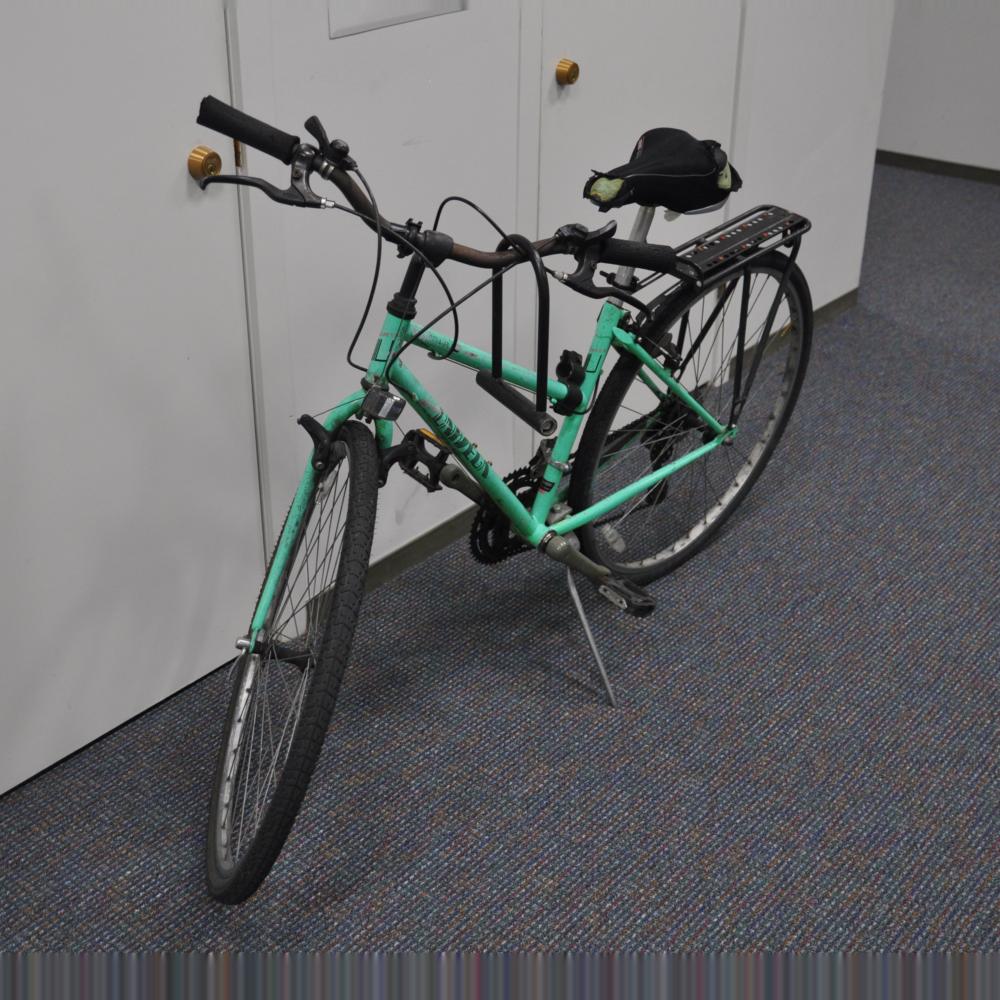}
  \includegraphics[width=0.23\linewidth]{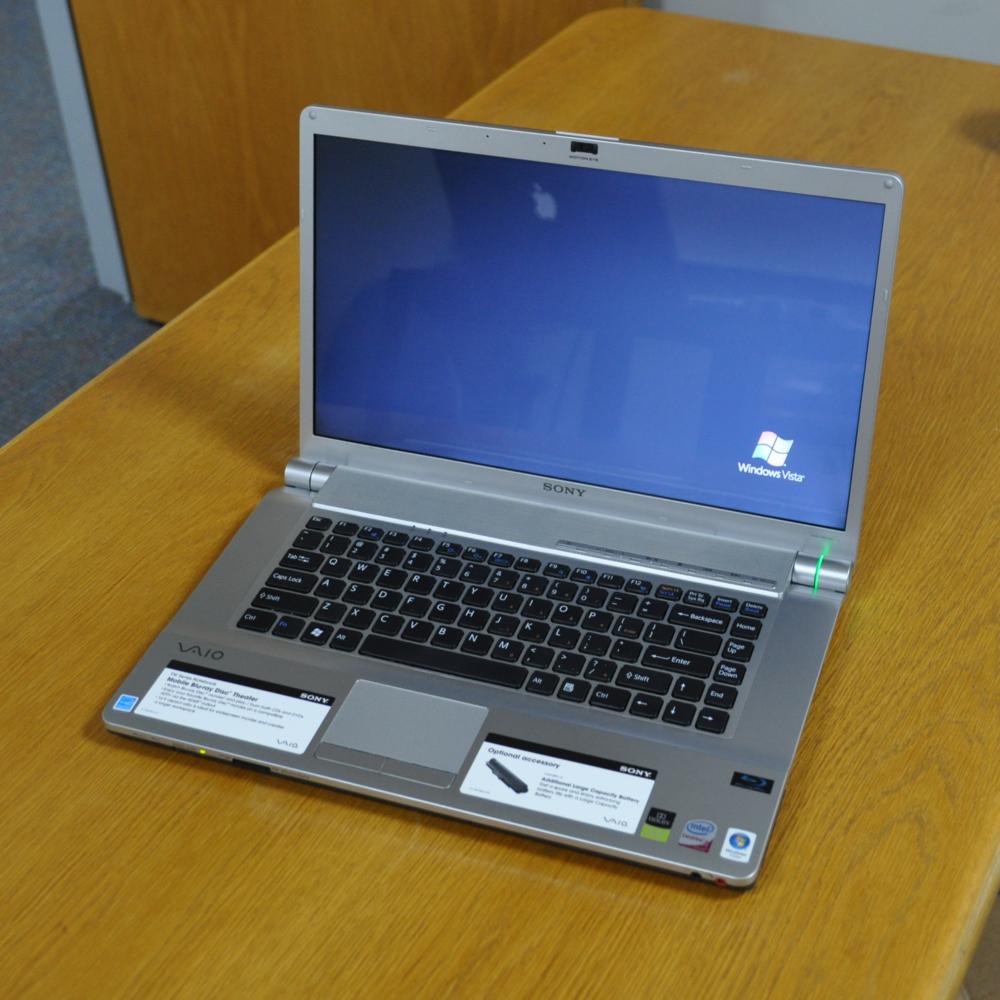}
  \includegraphics[width=0.23\linewidth]{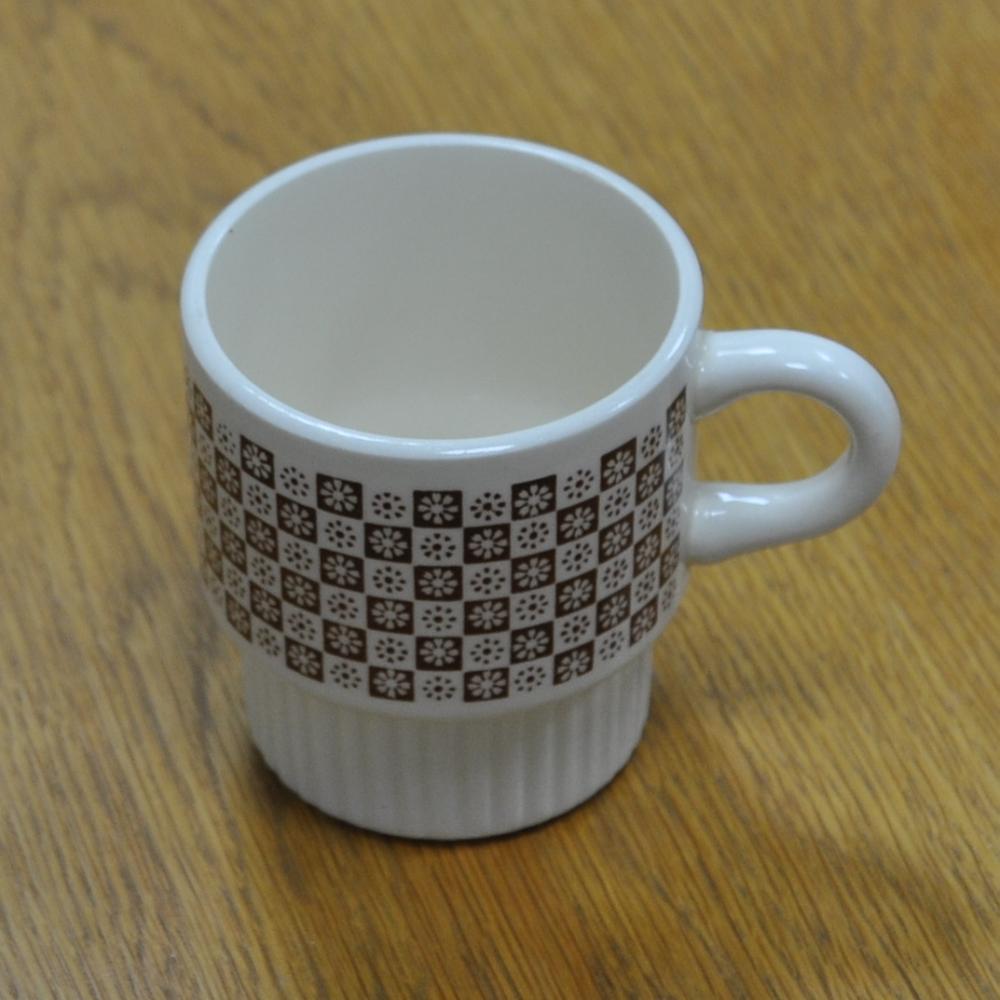}
  \caption{DSLR}
  \label{dslr}
  \end{subfigure}
  
    \begin{subfigure}[t]{0.49\linewidth}
  \centering
  \includegraphics[width=0.23\linewidth]{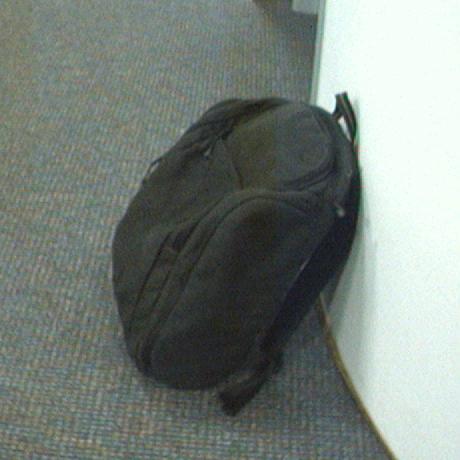}
  \includegraphics[width=0.23\linewidth]{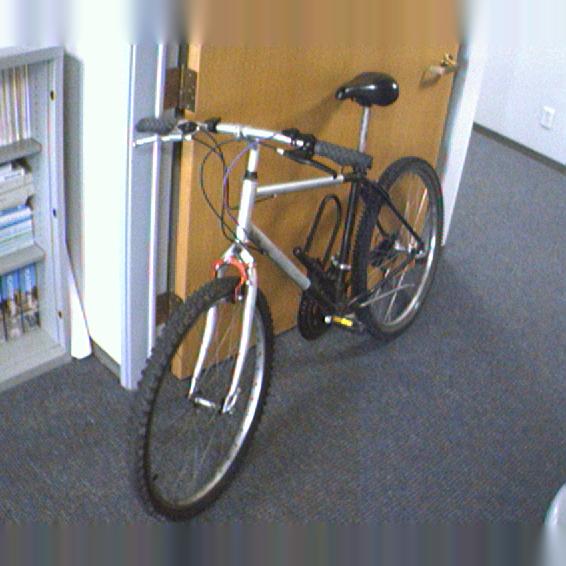}
  \includegraphics[width=0.23\linewidth]{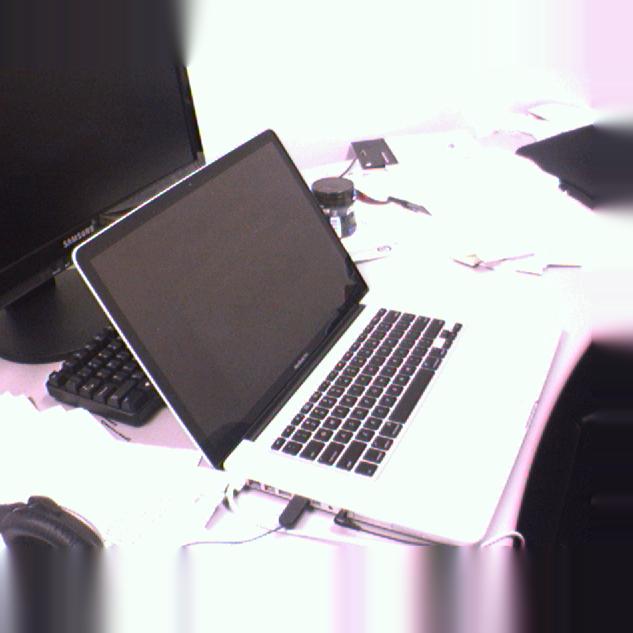}
  \includegraphics[width=0.23\linewidth]{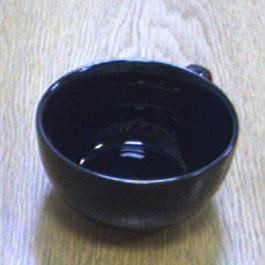}
  \caption{Webcam}
  \label{webcam}
  \end{subfigure}
    \begin{subfigure}[t]{0.49\linewidth}
  \centering
  \includegraphics[width=0.23\linewidth]{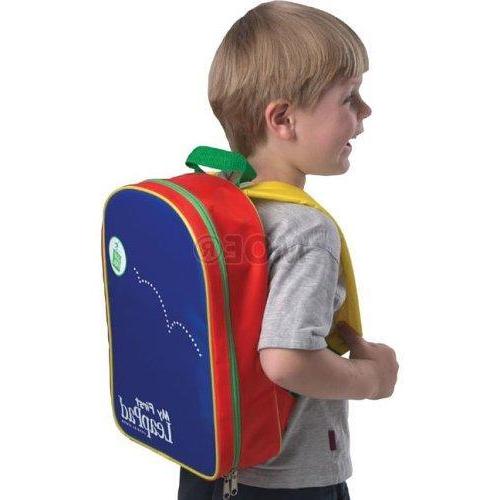}
  \includegraphics[width=0.23\linewidth]{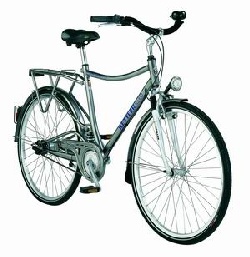}
  \includegraphics[width=0.23\linewidth]{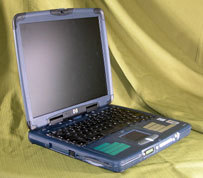}
  \includegraphics[width=0.23\linewidth]{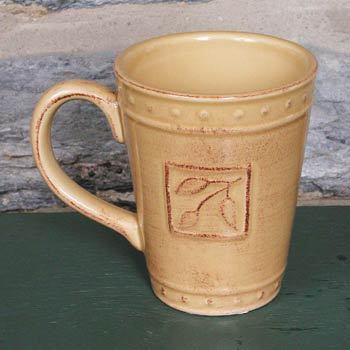}
  \caption{Caltech}
  \label{caltech}
  \end{subfigure}
\caption{\textbf{Example Images from the Office+Caltech Datasets.} Each set of images contains a random image with each of the labels: backpack, bike, laptop, and mug.}
\label{officecal}
\end{figure}

The Office+Caltech~\cite{saenko2010adapting} benchmark is formed from a set of 4 individual image datasets that share 10 categories and have SURF and DeCaf6 features available. The four domains that make up the Office+Caltech benchmark Amazon (A), Caltech (C), DSLR (D), and Webcam (W), are comprised of 958, 1123, 157, and 295 images, respectively. Example images from each domain and from four of the classes are shown in Figure~\ref{officecal}. Differences in pose, lighting, resolution, and background activity between the datasets are apparent in the images motivating the need for domain alignment. Each ordered pair of domains forms a separate experiment with a source and target domain for a total of 12 combinations. We use $X\rightarrow Y$ to denote aligning the source domain $X$ with target domain $Y$. Combined with the two separate feature sets for each domain, this provides 24 individual alignment experiments.

\begin{figure}
  \centering
  \begin{subfigure}[t]{0.4\linewidth}
  \centering
  \includegraphics[width=\linewidth]{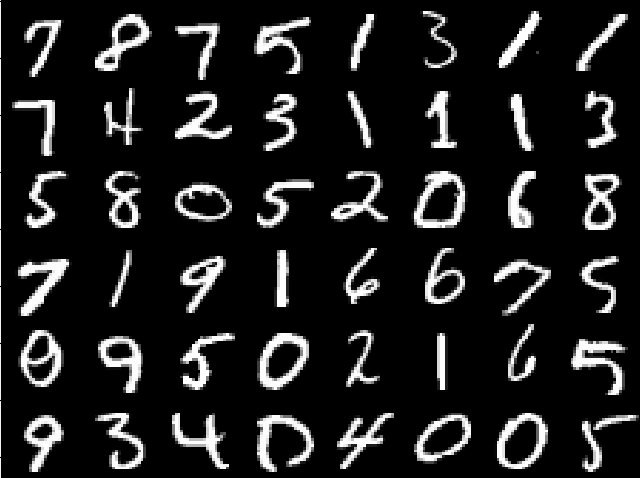}
    \caption{MNIST}
  \label{MNIST}
  \end{subfigure}
    \begin{subfigure}[t]{0.4\linewidth}
  \centering
  \includegraphics[width=\linewidth]{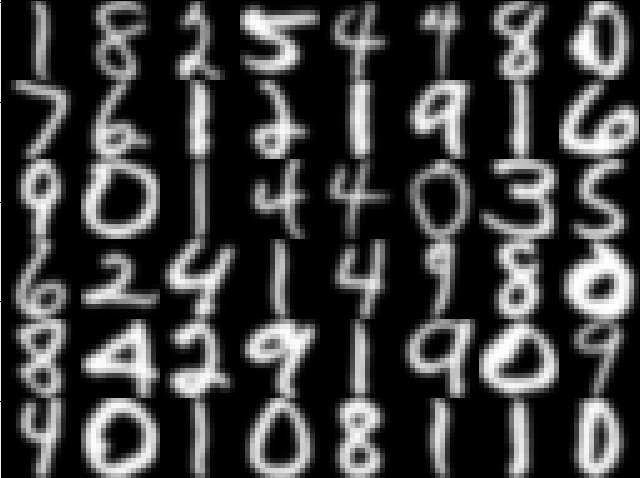}
    \caption{USPS}
  \label{USPS}
  \end{subfigure}
\caption{\textbf{Example from the MNIST and USPS datasets.} Each set of digits contains a random sample of 48 images. MNIST (\subref{MNIST}) images are 48x48 pixels. USPS (\subref{USPS}) images are 16x16 pixels.}
\label{mnistusps}
\end{figure}

The MNIST-USPS~\cite{long2013transfer} benchmark uses SURF features from a reduced set of 2000 images from the MNIST image dataset and 1800 from the USPS dataset. Each digit, 0-9, is represented in the images providing 10 class labels. A random sample of the digits in both datasets is given in Figure~\ref{mnistusps}. Using both MNIST and USPS as the source and target domains provides 2 separate experiments.

The final experiment uses 5 shared classes across the Caltech101~\cite{griffin2007caltech}, ImageNet~\cite{fang2013unbiased}, and VOC2007~\cite{everingham2010pascal} datasets. The domains contain the DeCaf6 features for $1415$, $7341$, and $3376$ images, respectively. The source/target combinations of the 3 domains provide 6 total experiments.

\subsection{Experimental Setup}
We follow the same experimental setup as previous work wherever possible and report the best performance for each method. In Office-Caltech, 800 SURF and 4096 DeCaf6 features are extracted from each image. In the MNIST-USPS experiment, 256 SURF features are calculated for each image. Caltech-ImageNet-VOC use 4096 DeCaf6 features. All features are normalized to have zero mean and unit variance. For each experiment, 20 randomized splits are created and results are given as the average across each split. The splits for the Office-Caltech datasets are the same splits as provided in 
\cite{gong2012geodesic}. In each experiment, the training set for the semi-supervised methods is composed of 20 labeled instances per class for the source domain (except for DSLR in which only 8 are used) and 3 labeled instances per class for the target domain. The target domain labels are used only for correspondences and not for training the final classifier. Additionally, we introduce a new domain and feature adaptation task using the Caltech-Office dataset to demonstrate FMA's applicability to heterogeneous domains. The source domain and target domains in this task differ in using SURF or DeCaf6 features.

We use the same hyper-parameters for both versions of filtered manifold alignment across both the Office-Caltech and the Caltech-ImageNet-VOC experiment sets. Nearest neighbor graphs are formed from instances within each dataset using cosine similarity. The edge weights for the 12 nearest neighbors of each sample are set to $\alpha*cos(v_i,v_j)$, where $\alpha=0.2$. Cross dataset correspondences between samples that share labels in the training set have edges weights of $1$. We calculate $20$ singular vectors from both the source and target domains and then combine them into a $n=40$ dimensional aligned embedding space. The edge weights and embedding dimensions were selected for high scores across the experiments. We report on the sensitivity of these parameters in section~\ref{subsec:fmaparams}. After alignment, a logistic classifier using categorical cross-entropy is trained on the labeled subset of the source domain and evaluated on the entire target domain. In the MNIST-USPS, experiments the number of nearest neighbors is reduced to 4 and the coefficient of the edge weights is decreased  to $\alpha=1.0$. All other parameters match those above.

The same parameters used in FMA are used for MA where applicable. Results for the unsupervised methods (MEDA, GFK, CORA) are reported from~
\cite{wang2018visual} for both the Office+Caltech and MNIST-USPS experiments. For the remaining methods, the default parameters given for each method are used. SSA requires the embedding size be less than or equal to the minimum number of labeled source samples for any class. To this end, we used an embedding dimension of 8 across most tasks and reduced it when necessary to facilitate convergence.

\subsection{Evaluation}
The classification accuracies for the Office-Caltech datasets using SURF and DeCaf6 features are presented in Tables~\ref{tab:surf} and~\ref{tab:decaf}, respectively. In addition, Table~\ref{tab:surf-decaf} presents the accuracies for the task where the source and target domains use different feature sets. The classification results for MNIST-USPS as well as Caltech-ImageNet-VOC are given in Table~\ref{tab:MU}. The best accuracies for each task are presented in boldface.

FMA demonstrated the second highest average accuracy across all the Office-Caltech experiments falling just below MEDA. Semi-supervised subspace alignment also has higher classification accuracy than FMA when using the SURF features, but does considerably worse with the Decaf6 features. Additionally, FMA had the highest classification accuracy on both the MNIST-USPS tasks and the Caltech-ImageNet-VOC tasks. These results demonstrate the effectiveness of filtered manifold alignment across a large range of different datasets and experiments. 

Table~\ref{tab:surf-decaf} shows the accuracy of FMA on the Office-Caltech experiments when the source and target domain have different feature sets. The accuracy on the target domains roughly matches those of the homogeneous domain tasks using the same features as the target domains. None of MEDA, GFK, CORAL, or SSA are applicable to domains with different feature sets.

\begin{table}
\centering
\caption{Office + Caltech 10 Classification Accuracy using SURF Features}
\label{tab:surf}
\bgroup
  \begin{tabular}{lrrrrr|rr}
  \toprule
     				& MEDA & GFK & CORAL & SSA & SMA  & FMA-I & FMA-F\\
    \cmidrule(r){2-8} 
    A$\rightarrow$C & 43.9 & 40.7 & \textbf{45.1} & 72.4 & 29.8 & 33.4 & 32.8\\
    A$\rightarrow$D & 45.9 & 40.1 &39.5& \textbf{58.3} & 56.8 & 56.9 & 57.4\\
    A$\rightarrow$W &  53.2 & 37.0 &44.4& 54.7 & \textbf{69.2} & 65.4 & 67.8\\
    C$\rightarrow$A & 56.5 & 46.0 &52.1& \textbf{59.8} & 47.9  & 49.2 & 48.1\\
    C$\rightarrow$D & 50.3 & 40.8 &45.9& \textbf{59.5} & 56.1 &  53.7 & 58.0\\
    C$\rightarrow$W & 53.9 & 37.0 &46.4& 54.0 & \textbf{68.7} &  65.3 & 67.2\\
    D$\rightarrow$A & 41.2 & 28.7 &37.7& \textbf{59.4} & 40.8 &  50.8 & 49.8\\
    D$\rightarrow$C & 34.9 & 29.3 &33.8& \textbf{71.2} & 27.0  & 34.8 & 33.2\\
    D$\rightarrow$W & \textbf{87.5} & 80.3 &84.7& 53.9 & 68.1  & 72.2 & 70.8\\
    W$\rightarrow$A & 42.7 & 27.6 &36.0& \textbf{60.1} & 45.5 &  50.0 & 50.3\\
    W$\rightarrow$C & 34.1 & 24.8 &33.7& \textbf{74.2} & 28.5  & 31.4 & 31.5\\
    W$\rightarrow$D & \textbf{88.5} & 85.4 &86.6& 54.3 & 56.5 & 56.1 & 59.0\\
    \cmidrule(r){1-8} 
    Ave & 52.7 & 43.1 & 48.8 & \textbf{61.0} & 49.6 &  51.6 & 52.2\\
    \bottomrule
  \end{tabular}
  \egroup
\end{table}

\begin{table}
\centering
\caption{Office + Caltech 10 Classification Accuracy using DeCaf6 Features}
\label{tab:decaf}
\bgroup
  \begin{tabular}{lrrrrr|rr}
  \toprule
     				& MEDA & GFK & CORAL & SSA & SMA & FMA-I & FMA-F\\
    \cmidrule(r){2-8} 
    A$\rightarrow$C & \textbf{87.4} & 79.2 &83.2& 75.9 & 69.2 & 84.6 & 85.8\\
    A$\rightarrow$D & 88.1 & 82.2 &84.1& 70.8 & 88.9 & \textbf{95.4} & 91.9\\
    A$\rightarrow$W & 88.1 & 70.9 &74.6& 85.6 & 88.7 & \textbf{95.0} & 94.5\\
    C$\rightarrow$A & \textbf{93.4} & 86.0 &92.0& 69.0 & 90.7 & 91.4 & 91.4\\
    C$\rightarrow$D & 91.1 & 86.6 &84.7& 59.8& 87.9 & \textbf{95.1} & 90.6\\
    C$\rightarrow$W & \textbf{95.6} & 77.6 &80.0&72.1& 89.1 & 92.5 & 92.0\\
    D$\rightarrow$A & \textbf{93.0} & 76.3 &85.5&70.0& 55.8 & 91.3 & 90.8\\
    D$\rightarrow$C & \textbf{87.5} & 71.4 &76.8&70.6& 48.4 & 85.6 & 85.7\\
    D$\rightarrow$W & 97.6 & \textbf{99.3} &\textbf{99.3}&72.5& 72.4 & 94.2 & 92.7\\
    W$\rightarrow$A & \textbf{99.4} & 76.8 &81.2&68.8& 66.7 & 91.6 & 91.9\\
    W$\rightarrow$C & \textbf{93.2} & 69.1 &75.5&72.4& 55.7 & 85.7 & 86.2\\
    W$\rightarrow$D & 99.4 & \textbf{100.0} &\textbf{100}&71.7& 86.1 & 96.7 & 91.9\\
    \cmidrule(r){1-8} 
    Ave & \textbf{92.9} & 74.0 &84.7&71.6& 74.6 & 91.6 & 90.5\\
    \bottomrule
  \end{tabular}
  \egroup
\end{table}

\begin{table}
\centering
\caption{Office 10 Classification Accuracy SURF to DeCaf6 Features}
\label{tab:surf-decaf}
\bgroup
  \begin{tabular}{crrrr}
  \toprule
    \multicolumn{1}{l}{} & \multicolumn{2}{c}{SURF to DeCaf6} & \multicolumn{2}{c}{DeCaf6 to SURF}\\ 
    & FMA-I & FMA-F & FMA-I & FMA-F\\
    \cmidrule{2-5}
    A$\rightarrow$C & $82.8$ & $82.9$ & $32.6$ & $33.4$\\
    A$\rightarrow$D & $90.2$ & $88.6$ & $58.3$ & $58.9$\\
    A$\rightarrow$W & $92.9$ & $94.4$ & $69.9$ & $70.0$\\
    C$\rightarrow$A & $89.8$ & $89.6$ & $49.9$ & $49.7$\\
    C$\rightarrow$D & $91.7$ & $89.6$ & $58.3$ & $58.4$\\
    C$\rightarrow$W & $90.7$ & $91.5$ & $68.4$ & $68.8$\\
    D$\rightarrow$A & $90.2$ & $90.1$ & $50.0$ & $49.8$\\
    D$\rightarrow$C & $84.1$ & $84.7$ & $32.4$ & $33.0$\\
    D$\rightarrow$W & $92.7$ & $93.5$ & $71.8$ & $70.8$\\
    W$\rightarrow$A & $90.8$ & $91.2$ & $50.0$ & $50.4$\\
    W$\rightarrow$C & $83.8$ & $84.6$ & $30.3$ & $31.7$\\
    W$\rightarrow$D & $92.7$ & $91.1$ & $60.1$ & $59.7$\\
    \cmidrule{1-5}
    Ave & $89.4$ & $89.3$ & $52.7$ & $52.8$\\
    \bottomrule
  \end{tabular}
  \egroup
\end{table}

\begin{table}
\centering
\caption{MNIST-USPS and Caltech-ImageNet-VOC Classification Accuracy}
\label{tab:MU}
\bgroup
\begin{tabular}{crrrrr|rr}
\toprule
& MEDA & GFK & CORAL & SSA & SMA  & FMA-I & FMA-F\\
\cmidrule(r){2-8}
M$\rightarrow$U & \textbf{89.5} & 31.2 & 49.2 &63.9 & 66.0 & 87.3 & 76.3\\
U$\rightarrow$M & 72.1 & 46.5 & 30.5 & 73.3 & 58.0 & \textbf{81.0} & 66.5 \\ 
\cmidrule{1-8}
Ave & 80.8 & 38.9 & 39.9 & 68.6 & 62.0 & \textbf{84.2} & 71.4\\
\bottomrule
\\
C$\rightarrow$I & 76.1 & 49.3 & 21.3 & 50.0 & 17.4 & \textbf{81.9} & 76.0\\
C$\rightarrow$V & 54.3 & 30.7 &48.6& \textbf{63.0}& 45.0 & 58.9 & 62.5\\
I$\rightarrow$C & 73.1 & 78.7 &35.5& 26.4& 61.6 & \textbf{99.8} & 99.6\\
I$\rightarrow$V & \textbf{67.3} & 64.8 &34.0& 65.0& 38.3 & 60.5 & 62.5\\
V$\rightarrow$C & 95.6 & 56.0 &69.6& 24.8& 67.4 & \textbf{99.8} & 93.3\\
V$\rightarrow$I & 74.7 & 61.3 &42.3& 50.7& 48.8 & \textbf{84.4} & 78.4\\

\cmidrule(r){1-8} 
Ave & 73.5 & 56.8 &41.9& 46.9 & 46.4 & \textbf{80.9} & 78.7\\
\bottomrule
  \end{tabular}
  \egroup
\end{table}

\begin{figure*}
  \centering
  \includegraphics[width=\linewidth]{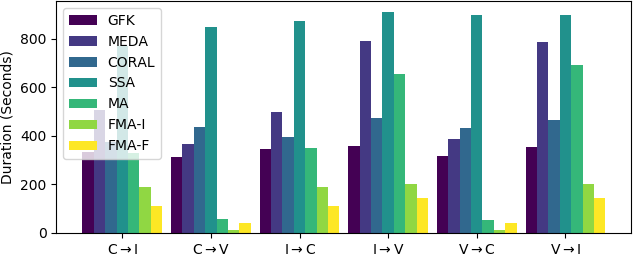}
  \caption{\textbf{Runtime comparison of alignment methods}. The time to align each pair of domains in the Caltech-ImageNet-VOC dataset is given for each of the 7 domain alignment methods tested.}
  \label{fig:runtime}
\end{figure*}

Figure~\ref{fig:runtime} shows the time in seconds to align each pair of domains in the Caltech-ImageNet-VOC experiments from a cold start. Each method was implemented in Python3.7.5 and ran on an 3.60GHz AMD Ryzen 7 3700X Processor with 16GB of RAM. Across every experiment, FMA and linear FMA provide significant time advantages over the other compared techniques. Additionally, experiments involving the larger ImageNet dataset as either the source or target domain demonstrate the superior complexity of feature-level FMA when the number of samples eclipses the number of features in the dataset as compared to instance-level FMA.

\subsection{Inductive Performance}
Using the MNIST-USPS~\cite{long2013transfer} benchmark, we evaluate the ability of feature-level filtered manifold alignment to generalize to unseen samples in the target domain. Samples that aren't initially available can't used for the graph construction portion of alignment. This can lead to forming a graph that is not an accurate representation of the domain's underlying manifold. However, unlike instance-level FMA, feature-level FMA learns a transformation matrix that can be applied to project new samples onto the joint manifold even though they weren't used in the original alignment phase. 

\begin{figure}
  \centering
  \begin{subfigure}[t]{0.49\linewidth}
  \centering
  \includegraphics[width=\linewidth]{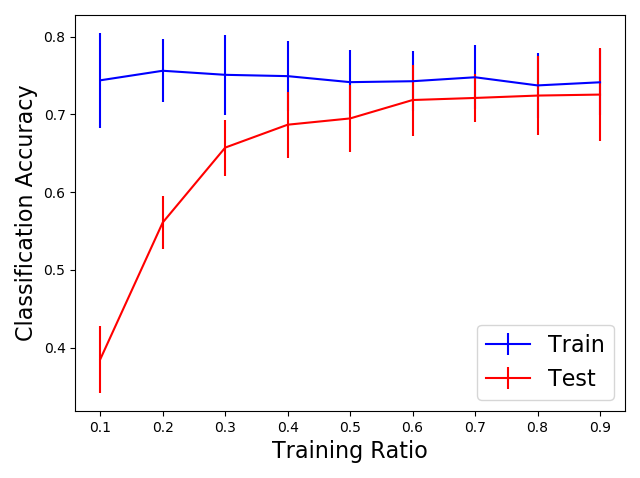}
  \caption{MNIST$\rightarrow$USPS}
  \label{uspsind}
  \end{subfigure}
  \begin{subfigure}[t]{0.49\linewidth}
  \centering
  \includegraphics[width=\linewidth]{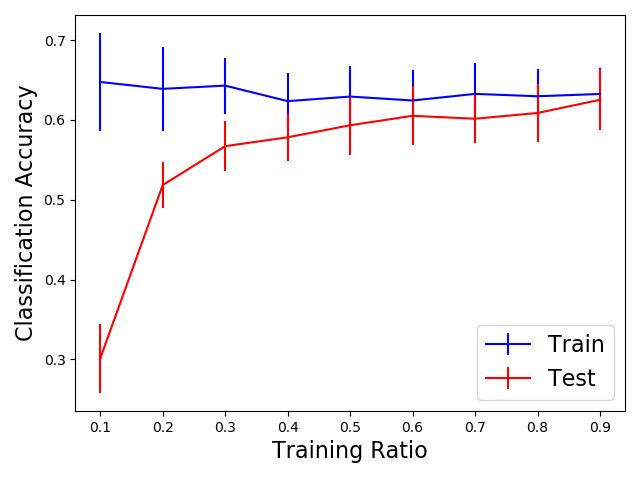}
  \caption{USPS$\rightarrow$MNIST}
  \label{mnistind}
  \end{subfigure}
\caption{\textbf{Inductive classification accuracy of FMA-F on MNIST-USPS} Classification accuracy on the training samples and the withheld test samples are given for USPS as the target domain (\subref{uspsind}) and for MNIST as the target (\subref{mnistind}). As the training size expands, the classification accuracy approaches the training accuracy.}
\label{fmaind}
\end{figure}

For the inductive experiment, the hyper-parameters and experimental setup are the same as the MNIST-USPS experiment above, but only a portion of the target domain samples are used in the alignment phase. The remaining portion are withheld and then embedded using the learned linear transform on the features. We vary the portion of the target domain used for training. Figure~\ref{fmaind} illustrates the results on both the MNIST to USPS transfer and the reverse. As the portion of the target domain samples used in training grows, the classification accuracy of the withheld samples approaches the accuracy of the training samples. Using $60\%$ or roughly 1200 samples from the target domain builds an accurate graph of the domain manifold so that the learned joint embedding and transformation can generalize to new samples with only a minor loss in classification accuracy.

\subsection{Hyper-Parameter Evaluation}
\label{subsec:fmaparams}
Using the office-Caltech dataset with SURF and DeCaf6 features, we study FMA's sensitivity to hyper-parameters for both instance-level and feature-level algorithms. In each experiment, we vary one parameter while keeping the others constant. The constant hyper-parameters match those used in the previous office-Caltech experiments: the edge weight coefficient is $\alpha=0.2$ and the embedding dimension is set to $40$.

\begin{figure*}
  \centering
  \includegraphics[width=\linewidth]{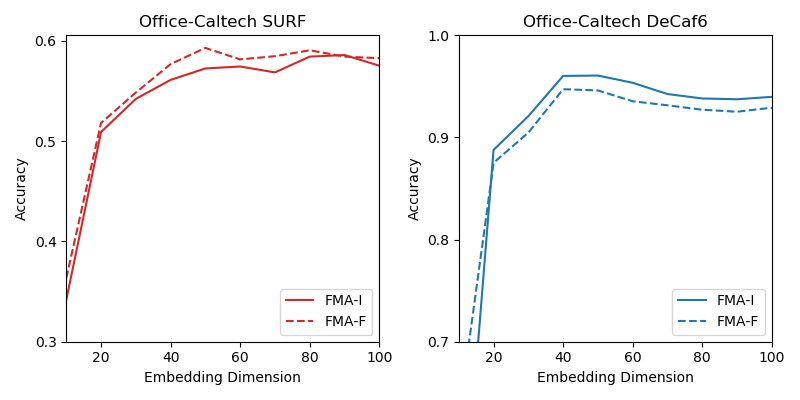}
  \caption{\textbf{Classification accuracy as a function of embedding dimension}. The effect on the overall classification accuracy for the office-Caltech dataset due to varying the final embedding dimension.}
  \label{fig:eigplot}
\end{figure*}

Figure~\ref{fig:eigplot} presents the overall classification accuracy for the office-Caltech experiment for different final embedding dimensions. The initial embedding size for each individual dataset is always half of the final embedding size in each experiment.

\begin{figure*}
  \centering
  \includegraphics[width=\linewidth]{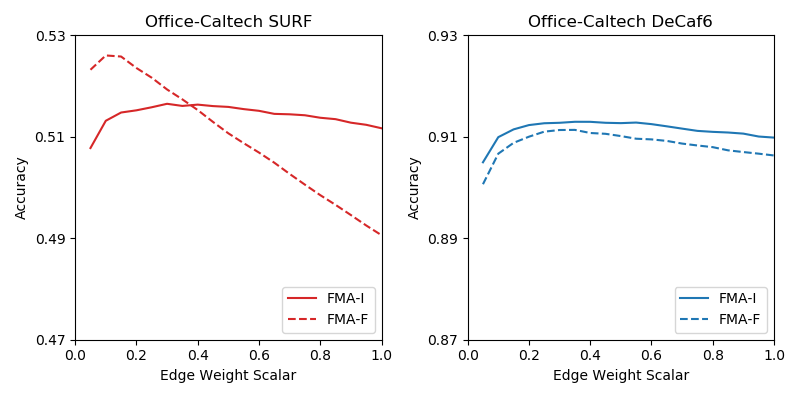}
  \caption{\textbf{Classification accuracy as a function of $\alpha$}. The effect on the overall classification accuracy for the office-Caltech dataset due to varying the edge weights of the nearest neighbor graphs.}
  \label{fig:alpha}
\end{figure*}

Figure~\ref{fig:alpha} demonstrates the effect of varying the scaling on the edge weights of the nearest neighbor graphs for each domain. A smaller scalar places less emphasis on the final embeddings of like instances within a domain having similar embeddings while a large value de-emphasizes the cross-domain correspondences in comparison. Note that increments on the y-axes are much smaller in comparison to Figure~\ref{fig:eigplot} due to scaling having a much smaller effect on the overall classification accuracy.

While the effect of the embedding dimension on the classification accuracy is large, it is due in part to the logistic classifier and the accuracy does somewhat level off for sufficiently large dimensions. Combined with the small effect of the edge weights on the final accuracy, this shows that FMA is not reliant on finding the best hyper-parameters and can perform well over a wide range of them.

\section{Conclusion}
\label{sec:conclusion}
Filtered manifold alignment provides a novel approach to performing manifold alignment and domain adaptation. Across a large number of experimental test beds, FMA meets or exceeds other state-of-the-art methods, while being considerably faster in calculating the alignment. Additionally, because we present both an instance-based and a feature-based version, the method can scale with the smaller of the two options. 

\section*{Acknowledgements}
This research was sponsored by the U.S. Army Research Laboratory and the U.K. Ministry of Defence under Agreement Number W911NF-16-3-0001. The views and conclusions contained in this document are those of the authors and should not be interpreted as representing the official policies, either expressed or implied, of the U.S. Army Research Laboratory, the U.S. Government, the U.K. Ministry of Defence or the U.K. Government. The U.S. and U.K. Governments are authorized to reproduce and distribute reprints for Government purposes notwithstanding any copyright notation hereon.

\bibliographystyle{plain}
\bibliography{bib}
\end{document}